\documentclass{article}

\usepackage[english]{babel}

\usepackage[letterpaper,top=2cm,bottom=2cm,left=3cm,right=3cm,marginparwidth=1.75cm]{geometry}

\usepackage{amsmath, amssymb}

\usepackage{booktabs}

\usepackage{graphicx}
\usepackage{caption}
\usepackage{subcaption}

\usepackage{authblk}

\usepackage[colorlinks=true, allcolors=blue]{hyperref}

\begin{document}

\title{PolarSym: Polar Geometry-aware Attention for CAD Floorplan Parsing}

\author[1,2]{Kerui Chen\thanks{Corresponding author. Email: chenke0616@163.com.}}
\author[1]{Yiqing Wang\thanks{Email: 16638710483@163.com.}}
\author[2]{Kangzhou Xin\thanks{Email: xinkangzhou96@gmail.com.}}
\author[2]{Qinghan Zhang\thanks{Email: zhangqinghan331@126.com.}}
\author[1,2]{Songyang Ding\thanks{Email: ding\_songyang@outlook.com.}}

\affil[1]{School of Computer and Information Engineering, Henan University of Economics and Law, Zhengzhou, China}
\affil[2]{Third Dimension (Henan) Software Technology, Zhengzhou, China}

\maketitle

\begin{abstract}
CAD plan parsing constitutes a fundamental task in Building Information Modeling (BIM), whose objective is to automatically recover architectural elements such as walls, doors, windows, and furniture from 2D engineering drawings. Although existing Transformer-based methods can capture global semantic dependencies through self-attention mechanisms, they primarily infer spatial relationships from semantic features without explicitly modeling the inherent geometric symmetry of building layouts. Consequently, these methods are vulnerable to incorrect correspondences when handling long-range matching and complex symmetric spatial configurations. To address these issues, this paper proposes a polar-coordinate geometry-aware attention framework named PolarSym for CAD plan parsing. The framework explicitly decouples architectural geometric relationships into two complementary factors—direction and distance—and models them separately. It enhances structural consistency through directional constraints and establishes long-range symmetric correspondences via distance constraints. Additionally, a dynamic gating mechanism is employed to achieve synergistic fusion of the two geometric information streams while preserving the core transformer architecture. This design significantly improves geometric modeling capabilities with minimal computational overhead. Experimental results on a public CAD plan parsing dataset demonstrate that PolarSym achieves a performance improvement of 1.73\% PQ, 1.54\% RQ, and 4.31\% mIoU over the reproduced SymPoint V2 baseline under identical training configurations. Furthermore, PolarSym exhibits faster convergence and more stable optimization dynamics. Ablation studies further validate the complementary role of directional and distance modeling. The results indicate that PolarSym effectively enhances the geometric perception capability of transformers with low computational overhead, providing an efficient geometric modeling solution for CAD plan parsing tasks.
\end{abstract}


\section{Introduction}

Computer-Aided Design (CAD) planar diagram parsing aims to automatically identify building components such as walls, windows, doors, and furniture from architectural drawings and reconstruct their structured semantic representations. As a foundational task for Building Information Modeling (BIM), digital twins, and indoor scene understanding, this problem holds significant application value in intelligent buildings, automated architectural design, and indoor navigation. With the advancement of deep learning, neural network-based methods have gradually replaced traditional rule-driven algorithms, substantially improving the automation level and reconstruction accuracy of CAD planar diagram parsing.

In recent years, Transformers have demonstrated superior modeling capabilities over convolutional neural networks in CAD planar diagram parsing tasks due to their global self-attention mechanisms. The SymPoint series represents a representative attempt to incorporate Point Transformer into architectural floor plan parsing, where point-wise feature interactions enable competitive instance-level parsing performance. However, existing Transformer methods primarily construct attention based on semantic similarity, while spatial geometric relationships are often implicitly represented via positional encoding without explicit modeling of architectural geometric structures. When confronted with repeated layouts, mirror structures, or long-range symmetric targets, attention mechanisms relying solely on semantic similarity tend to degenerate into local associations, making it difficult to establish stable long-range structural correspondences.

Architectural plans exhibit distinct geometric patterns that differ significantly from natural images, with numerous building components arranged in regular axisymmetric or repetitive distributions along the building's primary axis. The structural relationships of such components must satisfy not only semantic consistency but also be constrained by direction consistency (Direction Consistency) and radial correspondence (Radial Correspondence). Here, direction determines whether two components undergo the same symmetric transformation, while distance dictates their spatial correspondence. Current position encodings or distance biases typically employ unified Cartesian coordinates, mixing different geometric attributes or assuming monotonic relationships between distance and relevance. Such modeling approaches fail to accurately describe the prevalent long-range symmetric structures in architectural plans, thereby limiting Transformers' capability to model global geometric relationships.

\begin{figure}[htbp]
\centering
\includegraphics[width=0.50\linewidth]{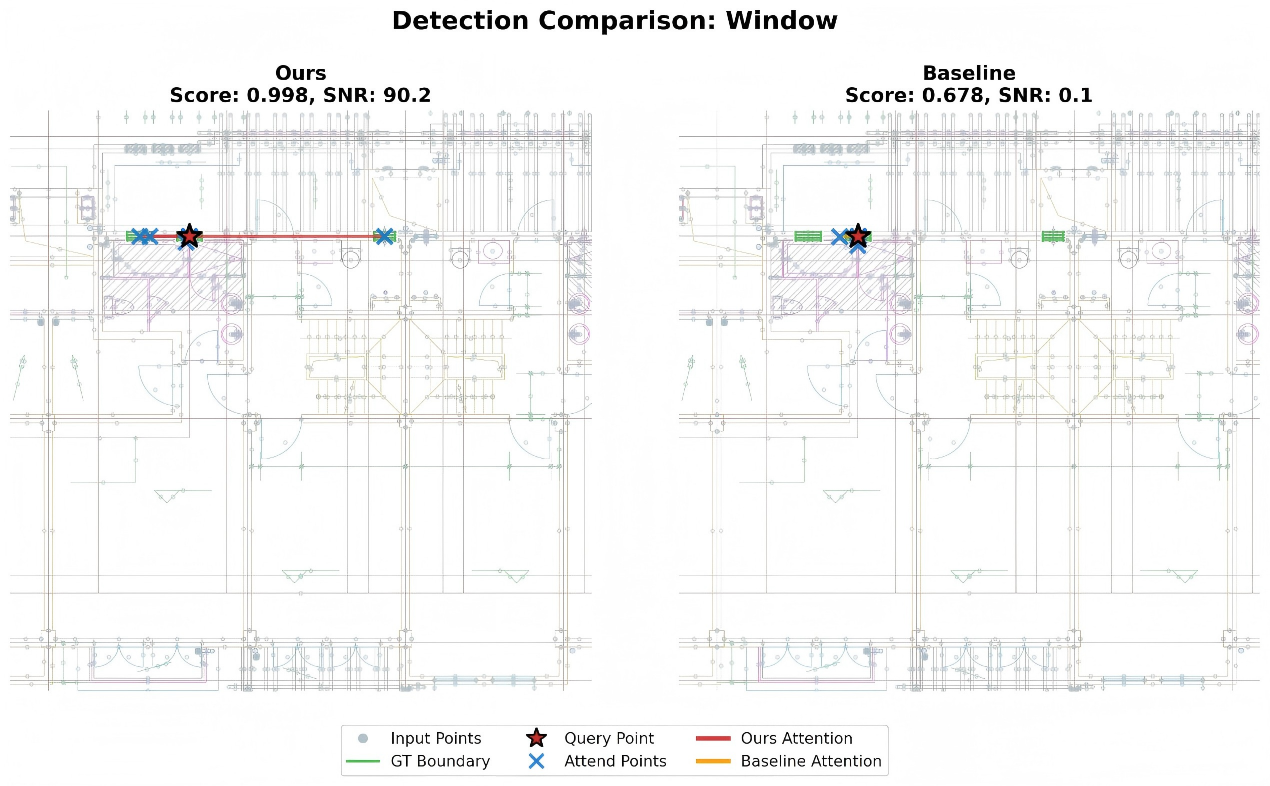}
\caption{Visual comparison of baseline models and our method.}
\label{fig:Fig1}
\end{figure}

To illustrate the aforementioned issue, Figure \ref{fig:Fig1} presents the attention visualization results of the baseline model and the proposed method under identical CAD inputs and query point (Query) conditions. It can be observed that the baseline model concentrates its attention primarily around the query point, establishing only local spatial associations while struggling to connect distant windows with identical structural attributes. In contrast, the proposed method not only accurately localizes the current target but also establishes long-range structural correspondences between multiple symmetric windows. This demonstrates that relying solely on semantic features is insufficient to fully capture the geometric patterns of component arrangement in architectural plane figures, whereas explicitly incorporating geometric constraints effectively enhances the Transformer's capability to model repetitive structures and symmetric relationships.

To address the aforementioned challenges, this paper proposes a polar-coordinate geometry-aware attention framework for CAD plane figure parsing—PolarSym. Unlike existing methods, PolarSym does not treat spatial information as a unified positional encoding. Instead, it explicitly decouples architectural geometric relationships during attention computation into two complementary factors—direction and distance—and models them separately. The direction branch enhances global directional consistency, while the distance branch establishes long-range structural correspondences. These two branches are jointly regulated by a unified dynamic gating mechanism to modulate the Transformer attention, thereby enabling synergistic semantic information and geometric prior modeling without altering the original network architecture.

The main contributions of this work are summarized as follows:
\begin{itemize}
    \item{Proposal of a polar-coordinate geometry-aware attention framework, PolarSym. Unlike conventional positional encodings, PolarSym explicitly models geometric relationships among architectural components in the attention layer by decoupling directional and distance information, providing a unified geometric modeling framework for CAD plane figure parsing.} 
    \item{Design of a direction–distance collaborative modeling mechanism. We propose a direction modeling method based on SF-RoPE (Split-Feature Rotary Position Embedding) and a radial modeling method based on RDB (Resonant Distance Bias), and leverage a dynamic gating strategy to achieve adaptive fusion of the two geometric modalities, thereby effectively enhancing the Transformer’s capability to represent long-range symmetric structures.} 
    \item{Empirical validation on public CAD plane figure parsing datasets. Under identical training configurations, PolarSym outperforms the SymPoint V2 baseline across multiple evaluation metrics including PQ, RQ, SQ, and mIoU, while demonstrating faster convergence and more stable optimization dynamics, thereby validating the effectiveness of explicit geometric modeling for architectural plane figure parsing tasks.} 
\end{itemize}

\section{Related Work}

\subsection{Planar CAD Diagram Parsing}

In recent years, the advancement of technologies such as Building Information Modeling (BIM), Digital Twin, intelligent construction, and indoor navigation has positioned automatic CAD diagram parsing as an emerging research direction within computer vision, graphics, and architectural intelligence. CAD diagrams not only encompass rich geometric primitives but also encode complex spatial topological relationships and architectural semantics, making the accurate modeling of spatial relationships between primitives a core challenge in CAD diagram understanding. Rezvanifar et al. conducted a systematic review of CAD diagram parsing methods, classifying existing approaches based on input data representation paradigms into four categories: Raster-based, Vector-based, Point-based, and Multimodal-based \cite{rezvanifar2019symbol}. Differing data representation paradigms determine the types of information models can leverage, thereby directly influencing the subsequent development trajectories of transformer architectures.

\subsubsection{Raster-based Methods}

Raster-based methods first convert CAD drawings into two-dimensional images before leveraging convolutional neural networks (CNNs) to learn visual features. Given that CNNs inherently possess local receptive fields and translation invariance, early architectural drawing recognition predominantly employed such approaches.

In recent years, multimodal fusion has further advanced raster-based methods. Xing et al. proposed a multimodal aggregation framework that represents architectural symbols as visual, geometric, and textual modalities \cite{xing2026multimodal}. This framework designs a hybrid sampling strategy while simultaneously employing boundary dense sampling (boundary dense sampling) and region-coarse sampling (region-coarse sampling) to extract symbol contour details and internal contextual information, respectively. Additionally, to enhance geometric representation capabilities, the method introduces normalized shape encoding (normalized shape encoding), which encodes geometric statistics—including area, perimeter, orientation angle, aspect ratio—into the feature space and integrates CAD textual information, thereby improving architectural symbol detection performance.

Although this approach effectively utilizes multimodal information, its underlying architecture remains within the CNN framework. Consequently, the model requires multiple rounds of sampling and feature fusion to acquire global contextual information, resulting in slow training convergence. Furthermore, the prevalence of repetitive elements in large-scale architectural drawings significantly burdens network computation, making it challenging to meet the demands of high-throughput automatic review in engineering scenarios.

\subsubsection{Vector-Based Methods}

Given that CAD inherently represents vector graphics, an increasing number of studies have directly leveraged CAD primitives for representation learning. Liu et al. first employed ResNet-152 to predict junction layers (Junction Layer) in architectural drawings, where topological nodes are recovered through intersection detection \cite{liu2017raster}. Subsequently, an image partition (IP) mechanism is utilized to reconstruct the primitive layer, encompassing primitives such as line segments, boundaries, and local contours. The junction layer fundamentally encodes node connectivity relationships within architectural spaces, while the primitive layer records geometric constituents of CAD drawings. Compared to direct pixel classification, this approach recovers structural information from architectural drawings. However, since IP primarily relies on regional segmentation and local structural constraints, it recovers only low-level geometric relationships, lacking higher-level topological representations between architectural components. Consequently, this method exhibits limited capability for long-range association modeling in complex architectural spaces.

Subsequently, Jiang et al. proposed YOLaT (You Only Look at Text), which first performed direct object detection on vector graphics \cite{jiang2021recognizing}. This method represents CAD primitives as Bezier curves and constructs a bi-stream graph neural network to jointly learn geometric and spatial relationships, enabling architectural component detection. Compared to raster-based methods, YOLaT avoids information loss caused by rasterization and directly leverages connectivity relationships between CAD primitives, thereby significantly improving detection accuracy. To further enhance vector graphics understanding, Dou et al. proposed YOLaT++, which not only expanded the VG-DCU large-scale vector graphics dataset but also established a comprehensive training dataset integrating vector graphics, rasterized images, annotations, and original CAD data, providing a unified foundation for transformer and graph network models \cite{dou2024hierarchically}. However, regardless of YOLaT or YOLaT++, both methods depend on Bezier curves to construct complex graph structures. As architectural scale increases, the number of Bezier curves grows exponentially, leading to heightened computational complexity in graph neural network message propagation and consequently reduced training stability and inference efficiency.

Concurrently, ReGroup employs recursive neural networks for hierarchical grouping of CAD primitives \cite{chaturvedi2021regroup}. The authors leverage contrastive learning to learn high-dimensional representations of primitive subsets and perform building component clustering based on cosine similarity. By eliminating the need for complex graph neural networks, ReGroup achieves substantially improved computational efficiency. However, this approach is limited to CAD drawings with simple hierarchical structures, no occlusions, and no layer overlaps—making it ineffective for handling complex spatial relationships in real-world architectural drawings.

PanCADNet introduces a CNN-GCN fusion network that jointly models geometric features (length, position), visual features, and type features (line segments, arcs) of CAD primitives \cite{fan2021floorplancad}. By leveraging graph convolutional networks (GCNs) to learn topological relationships between primitives, PanCADNet achieves unified identification of things and stuff. Nevertheless, the GCN-based graph construction and message-passing process incurs substantial computational overhead, particularly when architectural drawings contain numerous entities (e.g., complex commercial buildings), where the proliferation of nodes and edges may degrade both training and inference efficiency. Additionally, if annotations contain noise or entity boundaries are ambiguous, the GCN’s topological feature learning may be compromised.

Subsequently, GAT-CADNet further introduced relative spatial encoding (Relative Spatial Encoding, RSE) and cascaded edge encoding (Cascaded Edge Encoding, CEE) \cite{zheng2022gat}. Specifically, RSE establishes spatial encodings using relative coordinates of primitives, while CEE enhances information propagation within graph neural networks through multi-level edge representations, enabling joint optimization of semantic and instance recognition.

Following this, VectorFloorSeg proposed a bi-stream GNN attention network, where one stream learns line boundary information and the other stream learns regional information \cite{yang2023vectorfloorseg}. A cross-stream modulation mechanism was designed to facilitate inter-stream interactions, thereby achieving more complete planar floor segmentation. However, this approach exhibits several limitations: it is prone to misclassification in minimal regions (e.g., narrow corridors) due to insufficient sampling points leading to low feature embedding quality; it struggles to distinguish morphologically similar categories (e.g., wall panels versus railings), necessitating additional shape priors (e.g., window and door features); it currently only supports straight line segments, requiring polygonal approximation for complex curves (e.g., B-splines, Bezier curves), which may incur geometric precision loss; and the bi-stream graph neural network with cross-stream modulation increases model complexity, potentially resulting in slower inference speeds compared to lightweight image segmentation methods.

\subsubsection{Point-based Methods}

CADSpotting enables symbol recognition in large-scale architectural CAD drawings by representing each basic geometric shape through dense sampling points and describing its coordinate and color attributes \cite{yang2024cadspotting}. This approach addresses challenges such as symbol diversity, scale variations, and overlapping elements in CAD design, achieving integrated learning of semantic, instance, and panoptic segmentation. To achieve accurate segmentation in large and complex drawings, the authors further propose the Sliding Window Aggregation (SWA) technique, which combines weighted voting and non-maximum suppression (NMS). However, this method exhibits poor stability and lacks global topological characteristics.

SymPoint represents an initial attempt to address the panoptic symbol localization task in CAD drafting using point-set representations \cite{liu2024symbol}. Despite achieving significant success, SymPoint neglects geometric layer information and exhibits very slow convergence during training. To resolve these issues, SymPoint-V2 was introduced, building upon SymPoint by proposing a Layer Feature Enhancement Module (LFE) to encode geometric layer information into raw features and a Position-Guided Training (PGT) method to accelerate model convergence in early training stages \cite{liu2024sympoint}. In current benchmark tests, SymPoint-V2 demonstrates state-of-the-art performance among existing models.

\subsubsection{Multimodal-based Methods}

FloorNet employs a three-branch hybrid network that cross-fuses 3D point data and RGB data to fully leverage local spatial information and partial global layout information \cite{liu2018floornet}. This approach is computationally complex, exhibits low computational efficiency, and achieves low extraction accuracy.

Im2Vec is an end-to-end variational autoencoder (VAE) architecture where the encoder maps raster images to a global latent vector, and the decoder generates path latent vectors via recurrent neural networks (RNNs) \cite{reddy2021im2vec}. These path latent vectors are then decoded into closed Bézier curves through a path decoder, supporting variable path counts and topologies. However, this method relies on image-space losses that may cause fine-grained component loss, necessitating either increased training resolution or the design of more sophisticated loss functions for mitigation. Additionally, although it supports dynamic pruning of degenerate paths, the maximum path count $T$ must be pre-specified, limiting its capability for generating highly complex geometries.

RendNet constructs hypergraphs by converting vector graphics (VG) primitives (curves and surfaces) into hypergraphs, where nodes represent curve intersections or endpoints, edges correspond to curve segments, and hyperedges denote surfaces \cite{shi2022rendnet}. This architecture implements a dual-stream network: (i) a vector stream (hypergraph neural network (HGNN) for learning topological features) and (ii) a raster stream (via latent space rasterization (LSR), which renders VG into point clouds followed by PointNet for spatial feature extraction). The dual-stream features are fused through residual blocks before global aggregation to generate the target representation. Nevertheless, hypergraph construction relies on prior knowledge, with node selection (curve intersections and endpoints) and hyperedge definition (surface parameters) requiring manual design. Consequently, this approach may exhibit limited adaptability to complex or non-standard VG formats (e.g., custom curve types).

\subsection{Geometric Perception Attention}

In recent years, Transformer-based architectures have become a widely adopted solution in visual understanding tasks, benefiting from their ability to capture global spatial-contextual dependencies through self-attention mechanisms.

DeepSVG initially leveraged the hierarchical structure of transformers to learn vector graphic representation, partitioning SVG graphics into two levels: Shape-level and Command-level \cite{carlier2020deepsvg}. These levels encode high-level shape semantics and Bezier command sequences, respectively. The transformer first independently encodes each shape token before utilizing global self-attention to learn relationships between different shapes, thereby obtaining a global representation of the entire SVG graphic. DeepSVG fully exploits the arrangement invariance characteristic of SVG, enabling transformers to overcome the local receptive field limitations of CNNs and achieve generation and editing of complex vector graphics. However, this approach requires global attention computation over all shape tokens, resulting in a time complexity that becomes computationally prohibitive when CAD drawings contain tens of thousands of elements. Additionally, its spatial representation remains predominantly dependent on shape positions, lacking explicit modeling of directional relationships, distance constraints, and architectural topological constraints—thus hindering direct transfer to architectural CAD drawing parsing tasks.

Fan et al. proposed CADTransformer, decomposing architectural CAD drawings into primitive elements and leveraging HRNet to extract raster image features, which are then combined with element coordinates to generate primitive tokens \cite{fan2022cadtransformer}. To enhance local geometric relationship modeling, the authors designed neighborhood-aware self-attention (Neighborhood-aware Self-Attention), enabling information propagation from local to global scales through progressively expanding neighborhoods. They further introduced hierarchical feature fusion to integrate features across different scales and designed graphic entity position encoding for CAD element positioning. CADTransformer ultimately established Semantic Heads and Instance Heads to achieve architectural symbol category prediction and instance center offset prediction.

Compared to CNN and GCN methods, CADTransformer first demonstrated that transformers can effectively learn long-range contextual relationships within CAD drawings. However, its token features still rely on raster image features extracted by HRNet, fundamentally failing to escape raster representation. Additionally, the position encoding primarily uses absolute coordinate encoding, which only reflects element positions and cannot express the more critical directional and distance relationships between architectural components. When architectural drawings contain numerous repetitive room layouts, reliance on coordinate positions often leads to feature confusion across different instances.

The self-attention mechanism in transformers exhibits quadratic computational complexity, causing training and inference resource demands to escalate sharply as sequence length increases. To address this, researchers have proposed linear transformers that aim to improve attention computation efficiency. However, when sequence length exceeds the model dimension, linear transformers suffer from memory collisions, impairing precise retrieval of critical information. To resolve this, the Gated DeltaNet architecture was introduced, combining gating mechanisms with delta update rules to enhance linear transformers' performance in long-sequence modeling and information retrieval tasks \cite{yang2025gated}. Nevertheless, this method incurs additional storage overhead and has not been experimentally validated on vector graphics data, making it incapable of capturing fine-grained topological relationships within diagrams.

Representative methods such as the SymPoint series have introduced Point Transformers into CAD planar drawing parsing, significantly improving instance-level recognition performance. By representing architectural primitives as point entities, these methods effectively capture semantic interactions between different building components and achieve state-of-the-art performance on public benchmarks.

Despite these advances, existing transformer-based methods still primarily construct attention based on feature similarity. The geometric regularity of architectural layouts is implicitly encoded through positional embeddings, making it challenging to establish reliable correspondences between components that are structurally symmetric but distant. Consequently, parsing performance degrades in scenarios involving repetitive layouts, mirror structures, and long-range architectural dependencies.

To enhance spatial representation learning, numerous studies have incorporated geometric information into transformer attention mechanisms. Position encoding is the most widely adopted strategy, including absolute position encoding, relative position encoding, and rotational position embedding (RoPE) \cite{su2024roformer}. RoPE preserves relative positional relationships through feature rotation and has demonstrated exceptional performance in visual transformers and large language models. Beyond position encoding, some research integrates pairwise spatial information into attention computation to introduce geometric perception attention. Typical approaches employ relative positional offsets or distance-aware attention to improve local geometric consistency. These methods have achieved promising results in point cloud understanding, object detection, and 3D scene analysis.

However, existing geometric perception attention mechanisms remain insufficient for architectural planar drawing parsing. First, most position encoding methods represent geometry in Cartesian coordinates, embedding both directional and distance information into a single representation. This implicit encoding hinders the network's ability to distinguish distinct geometric factors during attention learning. Second, conventional distance-aware attention typically assumes a monotonic relationship between spatial distance and feature correlation. While this assumption holds for natural images and point clouds, it conflicts with architectural drawings, where symmetric structures are often separated by larger spatial distances. Consequently, traditional geometric attention mechanisms tend to underestimate long-range symmetric correspondences.

\subsection{Discussion}

The aforementioned review demonstrates that current CAD planar diagram parsing methods achieve robust semantic representations via Transformer architectures, while existing geometry-aware attention mechanisms enhance spatial modeling through positional or distance constraints. However, neither direction nor distance has been explicitly modeled as independent geometric priors, and the intrinsic symmetry of architectural layouts remains underutilized.

Inspired by this, we propose PolarSym, a polar coordinate geometry-aware attention framework that explicitly decomposes architectural geometry into two complementary components: direction and distance. Directional consistency is modeled via Split Feature Rotation Position Embedding (SF-RoPE), while long-range radial correspondences are captured by Resonance Distance Deviation (RDB). Combined with a dynamic gating mechanism, the proposed framework integrates geometric priors directly into Transformer attention, thereby enabling more reliable structural correspondences for CAD planar diagram parsing.

\section{Methods}

\subsection{Overall Framework}

\begin{figure}[htbp]

\centering
\includegraphics[width=0.85\linewidth]{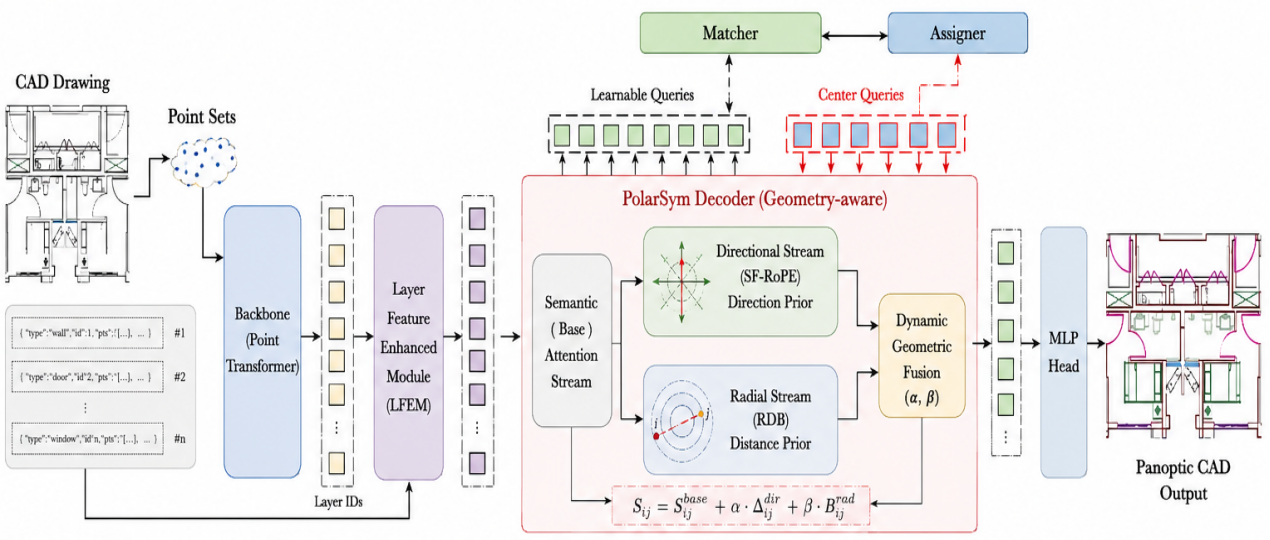}
\caption{PolarSym framework diagram.}
\label{fig:Fig2}
\end{figure}

The overall architecture of the proposed PolarSym framework is illustrated in Figure \ref{fig:Fig2}.

Given a set of point features extracted by the Point Transformer backbone, our objective is to enhance the attention mechanism through explicit geometric reasoning while preserving the original Transformer architecture.

Let $\mathbf{F} = \{f_i\}_{i=1}^N \in \mathbb{R}^{N \times D}$ represent the semantic feature embeddings of $N$ points, where $D$ denotes the feature dimension. Each point is associated with two-dimensional spatial coordinates. The data $\mathbf{P} = \{p_i = (x_i, y_i)\}_{i=1}^N \in \mathbb{R}^{N \times 2}$ is first fed into the backbone network, which generates query, key, and value representations via linear projection: $\mathbf{Q} = \mathbf{F} \mathbf{W}_Q,\quad \mathbf{K} = \mathbf{F} \mathbf{W}_K,\quad \mathbf{V} = \mathbf{F} \mathbf{W}_V$.

Subsequently, the layer ID features are fused into the Layer Feature Enhanced Module (LFEM), followed by the PolarSym decoder module. PolarSym does not compute attention scores based on semantic similarity directly but instead enhances attention scores using two complementary geometric priors. Specifically, the proposed framework explicitly decomposes geometric relationships into direction and distance, and models them through two parallel branches: Direction branch: Responsible for capturing global directional consistency via Split Feature Rotation Position Embedding (SF-RoPE); Radial branch: Responsible for modeling long-range geometric correspondences via Resonance Distance Bias (RDB).

The outputs of the two branches are integrated into the original attention scores through a lightweight dynamic gating mechanism to produce the final geometric-aware attention map. Therefore, the proposed framework retains the Transformer’s global semantic modeling capability while explicitly incorporating architectural geometric priors into the attention computation. Unlike existing approaches that implicitly embed spatial information into feature representations via positional encodings, PolarSym directly models geometry in the attention domain. This decoupled formulation enables the collaborative optimization of semantic representations, directional consistency, and radial correspondences without modifying the backbone architecture.

\subsection{PolarSym Geometric Attention}

Transformer attention estimates the correlation between two points based on semantic similarity:
\begin{equation}
S^{\mathrm{sem}} = \frac{\mathbf{Q} \mathbf{K}^\top}{\sqrt{d}}
\label{eq:semantic_attention}
\end{equation}
where $d$ denotes the feature dimension per attention head.

Although semantic attention effectively captures contextual relationships, it does not explicitly encode geometric regularities inherent in building layouts. In CAD floor plans, distant building elements may exhibit weak semantic relationships while still belonging to the same underlying geometric symmetry pattern. Consequently, relying solely on semantic similarity often leads to inaccurate long-range correspondences.

\begin{figure}[htbp]
\centering
\includegraphics[width=0.70\linewidth]{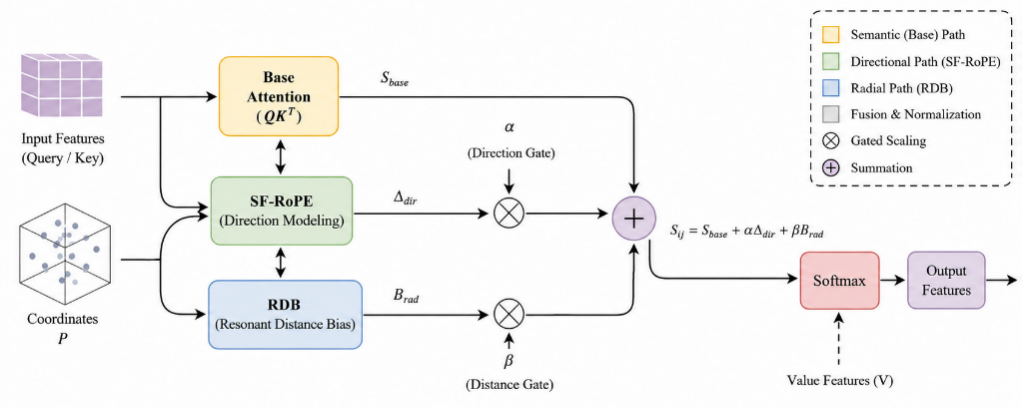}
\caption{PolarSym geometric attention computation.}
\label{fig:Fig3}
\end{figure}

To address this, we propose PolarSym geometric attention, which enhances semantic attention through explicit geometric reasoning. As shown in Figure \ref{fig:Fig3}.

PolarSym geometric attention does not represent geometry as a unified positional embedding, but decomposes geometric information into two complementary components: Direction consistency: describes whether two points satisfy the same symmetric transformation; Radial correspondence: quantifies whether two points exhibit compatible geometric distances under building symmetry.

Compared with traditional geometric perception attention, PolarSym introduces two critical distinctions. First, direction and radial information are independently modeled rather than jointly encoded into positional embeddings, enabling each geometric factor to focus on specific aspects of building component symmetry within the planar graph. Second, geometric information is injected via additive attention modulation rather than feature transformation, facilitating seamless integration into existing Transformer architectures with negligible computational overhead.

\subsubsection{Directional Modeling}

The directional branch aims to enhance attention through explicit orientation awareness. In architectural floor plans, symmetric structures typically exhibit consistent directional relationships relative to the building’s primary axis. However, conventional positional encoding implicitly preserves relative positions without explicitly distinguishing distinct symmetric orientations.

\begin{figure}[htbp]
\centering
\includegraphics[width=0.50\linewidth]{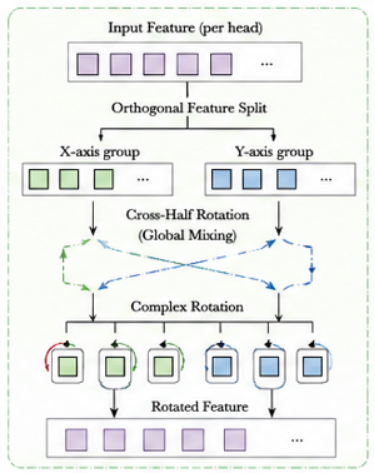}
\caption{Split feature rotation position embedding representation.}
\label{fig:Fig4}
\end{figure}

To address this issue, we propose Split Feature Rotation Position Embedding (SF-RoPE). As illustrated in Figure \ref{fig:Fig4}.

Unlike conventional Rotation Position Embedding (RoPE), which performs rotation on adjacent feature pairs, SF-RoPE divides feature dimensions into two complementary halves and executes cross-half feature rotation. This design enables long-range interactions between distant feature channels, thereby improving global directional representation.

Given a query (or key) feature vector $\boldsymbol{x}=[x^{(1)},x^{(2)}]$, SF-RoPE first splits the feature into two equal parts: $x^{(1)}=x_{1:D/2},\;x^{(2)}=x_{D/2+1:D}$. For each feature pair, its rotated representation is computed as:
\begin{equation}
\begin{bmatrix}
x'_k \\
x'_{k+D/2}
\end{bmatrix}
=
\begin{bmatrix}
\cos\theta_k & -\sin\theta_k \\
\sin\theta_k & \cos\theta_k
\end{bmatrix}
\begin{bmatrix}
x_k \\
x_{k+D/2}
\end{bmatrix}.
\label{eq:sf_rope_rot}
\end{equation}
where the rotation angle $\theta_k$ is determined by physical coordinates: $\theta_k=\lambda \phi(\boldsymbol{p}_i)$. Here, $\phi(\boldsymbol{p})$ denotes an angle mapping function that takes positional coordinates as input, and $\lambda$ is a learnable frequency scaling factor. Different attention heads alternately adopt horizontal and vertical coordinates to encode directional information along the two primary axes.

The rotated query and key are then used to compute directional attention: $S^{\mathrm{dir}}=\frac{\mathbf{Q}'\mathbf{K}'^\top}{\sqrt{d}}$. The directional enhancement term is obtained via:
\begin{equation}
\Delta^{\mathrm{dir}} = S^{\mathrm{dir}} - S^{\mathrm{sem}}
\label{eq:dir_enhance}
\end{equation}

The directional branch does not replace semantic attention but solely estimates geometric-induced attention increments. Consequently, semantic representations and directional geometry remain decoupled, enabling the network to explicitly learn architectural symmetry while preserving original semantic relationships.

In contrast to conventional RoPE, which establishes local adjacent channel rotations, SF-RoPE facilitates global channel interactions. This design enables more effective modeling of long-range directional consistency prevalent in CAD floor plans.

\subsubsection{Radial Modeling}

Although directional branches can establish global directional consistency, relying solely on directional information is insufficient to determine whether two points exhibit reasonable spatial correspondence. Therefore, we further introduce a radial branch to model pairwise distance relationships, thereby enhancing long-range geometric correlation capabilities.

Existing studies demonstrate that directly incorporating spatial distance information into attention computation can effectively improve geometric modeling capabilities, such as through distance bias or relative spatial encoding modulation of attention weights. However, such methods generally perform well in local spatial modeling, while symmetric components in architectural floor plans often span extensive spatial ranges. Sole reliance on local distance constraints hinders the accurate establishment of long-range structural correspondences.

\begin{figure}[htbp]
\centering
\includegraphics[width=0.70\linewidth]{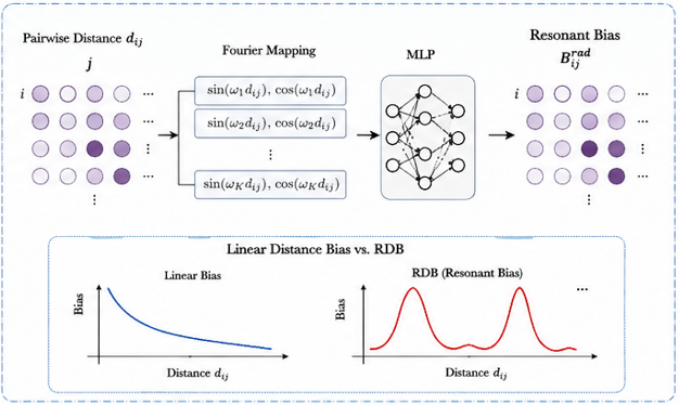}
\caption{Computation process of resonant distance bias.}
\label{fig:Fig5}
\end{figure}

To address this, we propose the Resonant Distance Bias (RDB) module. This module learns a non-monotonic response function in the distance space to assign higher attention responses to distant point pairs with structural correspondence, without assuming a fixed decay relationship between distance and correlation. As illustrated in Figure \ref{fig:Fig5}.

For two points located at $\boldsymbol{p}_i=(x_i,y_i)$ and $\boldsymbol{p}_j=(x_j,y_j)$, their Euclidean distance is first computed: $\boldsymbol{d}_{ij} = \|\boldsymbol{p}_i - \boldsymbol{p}_j\|_2$. The RDB module does not directly utilize the scalar distance $d$ but transforms it into a Fourier feature representation:
\begin{equation}
\gamma(d_{ij}) = \bigl[d_{ij},\sin(2^0 \pi d_{ij}),\cos(2^0 \pi d_{ij}),\dots,\sin(2^{L-1} \pi d_{ij}),\cos(2^{L-1} \pi d_{ij})\bigr]
\label{eq:fourier_gamma}
\end{equation}
where $L$ denotes the number of frequency bands. The encoded distance features are then processed by a lightweight multilayer perceptron (MLP):
\begin{equation}
B^{\mathrm{rad}} = \mathrm{MLP}\bigl(\gamma(d_{ij})\bigr)
\label{eq:radial_bias}
\end{equation}
This network predicts the radial attention bias.

Unlike linear distance bias, Fourier encoding enables the network to learn periodic distance responses with multiple local maxima. Consequently, the radial branch assigns higher attention values to distant but symmetric structures while suppressing geometrically inconsistent point pairs. This resonant mechanism significantly improves the establishment of long-range architectural component correspondences without introducing artificial distance priors.

\subsubsection{Dynamic Gating Strategy}

The directional branch and radial branch capture complementary geometric properties but exhibit distinct optimization characteristics. Directional consistency represents a globally stable structural prior throughout training, whereas radial relationships depend on the learning of nonlinear distance functions, rendering them more sensitive to parameter initialization. To balance their optimization processes, PolarSym introduces a lightweight dynamic gating strategy that adaptively controls the contribution of each geometric branch.

Consequently, the final attention score is formulated as:
\begin{equation}
\boldsymbol{S} = S^{\mathrm{sem}} + \alpha \Delta^{\mathrm{dir}} + \beta B^{\mathrm{rad}}
\label{eq:final_attn_score}
\end{equation}
where $S^{\mathrm{sem}}$ denotes the original semantic attention, $\Delta^{\mathrm{dir}}$ represents the directional enhancement term generated by the SF-RoPE branch, $B^{\mathrm{rad}}$ corresponds to the radial distance deviation learned by the RDB branch, and $\alpha$, $\beta$ are the physically interpretable learnable gating coefficients that control the contributions of the two geometric branches.

To stabilize optimization, both gating coefficients are parameterized as: $\alpha=\tanh(g_\alpha)$, $\beta=\tanh(g_\beta)$, where $g_\alpha$ and $g_\beta$ are trainable parameters.

The hyperbolic tangent function constrains the coefficients within the interval $(-1,1)$, preventing excessive geometric responses during early optimization. Experiments initially adopted asymmetric initialization: the directional gate was set to a larger value $g_\alpha=2.0$, $\alpha_{\mathrm{init}}=\tanh(2.0)\approx0.96$ to ensure directional geometry participates in attention computation from the outset. The radial gate was initialized to $g_\beta=0$, $\beta_{\mathrm{init}}=\tanh(0)=0$, suppressing the radial branch during the initial phase, while its internal parameters were optimized via backpropagation. As training progressed, the network gradually increased $\beta$ to incorporate radial geometry only after establishing reliable semantic and directional representations.

\section{Experiments}

\subsection{Experimental Setup}

The PolarSym framework was evaluated on publicly available CAD planar diagram parsing benchmarks. In this work, four standard metrics for panoptic segmentation and semantic segmentation tasks were employed: Panoptic Quality (PQ), Recognition Quality (RQ), Segmentation Quality (SQ), and Mean Intersection over Union (mIoU). The core notations are defined as follows: TP denotes the number of instances where predicted results successfully match ground-truth labels (true positives); FP denotes the number of predicted instances without corresponding ground-truth label matches (false positives); FN denotes the number of ground-truth instances not predicted by the model (false negatives); $\text{IoU}_i$ denotes the intersection-over-union (IoU) for matched pairs.

SQ quantifies the pixel-level segmentation accuracy of matched instances and is defined as the mean intersection-over-union (IoU) across all successfully matched instances, calculated as:
\begin{equation}
\mathrm{SQ} = \frac{\sum\limits_{i\in\mathrm{TP}} \mathrm{IoU}_i}{|\mathrm{TP}|}
\label{eq:SQ}
\end{equation}

RQ measures the completeness of instance-level recognition and matching, equivalent to a weighted F1 score, effectively characterizing the model’s capability for target instance detection and recognition. Its calculation formula is:
\begin{equation}
\mathrm{RQ} = \frac{|\mathrm{TP}|}{|\mathrm{TP}|+\frac{1}{2}|\mathrm{FP}|+\frac{1}{2}|\mathrm{FN}|}
\label{eq:RQ}
\end{equation}

PQ serves as a comprehensive evaluation metric for panoptic segmentation, integrating both pixel-level segmentation quality and instance-level recognition quality. It is derived from the product of SQ and RQ, with the overall calculation formula expressed as:
\begin{equation}
\mathrm{PQ} = \mathrm{SQ} \times \mathrm{RQ} = \frac{\sum\limits_{i\in\mathrm{TP}} \mathrm{IoU}_i}{|\mathrm{TP}|+\frac{1}{2}|\mathrm{FP}|+\frac{1}{2}|\mathrm{FN}|}
\label{eq:PQ}
\end{equation}

mIoU is a classical evaluation metric for semantic segmentation, computed as the average intersection-over-union (IoU) across all classes to holistically assess the model’s global segmentation performance. Given $N_c$ as the total number of classes in the dataset and $\mathrm{IoU}_c$ as the intersection-over-union for class $c$, the calculation formula is:
\begin{equation}
\mathrm{IoU}_c = \frac{\mathrm{TP}_c}{\mathrm{TP}_c+\mathrm{FP}_c+\mathrm{FN}_c},\quad \mathrm{mIoU} = \frac{1}{N_c}\sum_{c=1}^{N_c}\mathrm{IoU}_c
\label{eq:mIoU}
\end{equation}

All experiments were conducted on two NVIDIA V100-32GB GPUs using identical training configurations. Following the constrained resource setup adopted in this work, the network was trained for 50 epochs using the AdamW optimizer.

\subsection{Comparison with State-of-the-Art Methods}

Table \ref{tab:sota_comparison} compares PolarSym with representative planar graph parsing methods. Although PolarSym achieves competitive performance with a significantly smaller computational budget than previous state-of-the-art methods, it still demonstrates superior results. Specifically, PolarSym attains 87.15\% PQ, significantly outperforming PanCADNet, GAT-CADNet, and SymPoint V1. Compared to the official report of SymPoint V2, our model achieves comparable performance in just 50 training epochs, whereas SymPoint V2 was trained for 250 epochs under conditions of 8×GPUs and batch size 16, while SymPoint V1 required up to 1,000 epochs. These results demonstrate that explicit geometric modeling significantly enhances optimization efficiency. PolarSym does not rely solely on prolonged optimization but enables the Transformer to establish reliable structural correspondences in early training stages, thereby improving data utilization efficiency within constrained computational resources.

To eliminate the influence of varying training budgets, we further compare PolarSym with the re-implemented SymPoint V2 benchmark under identical constrained settings. As shown in Table \ref{tab:sympoint_compare}, PolarSym consistently outperforms the re-implemented benchmark across all evaluation metrics, with improvements of +1.73\% in PQ, +1.54\% in RQ, +0.30\% in SQ, and +4.31\% in mIoU.

The most significant improvement is observed in mIoU, indicating that the proposed geometry-aware attention mechanism substantially enhances semantic consistency. Furthermore, the improvement in RQ demonstrates that explicit geometric reasoning facilitates the establishment of more reliable long-range correspondences, thereby improving instance recognition.

Overall, these results demonstrate that incorporating explicit directional and radial geometric priors into the Transformer attention mechanism can effectively improve planar graph parsing without increasing network complexity.

\begin{table}[htbp]
\centering
\caption{Comparison between PolarSym and representative planar graph parsing methods.}
\label{tab:sota_comparison}
\resizebox{\linewidth}{!}{
\begin{tabular}{l p{2.8cm} c c c c c}
\toprule
Method & Backbone & Epochs & PQ (\%) & RQ (\%) & SQ (\%) & mIoU \\
\midrule
PanCADNet & CNN + GCN & $-$ & 59.5 & 82.6 & 66.9 & $-$ \\
GAT-CADNet & GAT (Graph Attention Network) & $-$ & 73.7 & 91.4 & 80.7 & $-$ \\
SymPoint V1 & PointNet++ & 1000 & 83.3 & 91.4 & 91.1 & 69.7 \\
SymPoint V2 & PointT & 250 & 90.1 & 96.3 & 93.6 & 74.0 \\
PolarSym (Ours) & PointT & 50 & \textbf{87.15} & \textbf{91.48} & \textbf{95.28} & \textbf{70.87} \\
\bottomrule
\end{tabular}
}
\end{table}

\begin{table}[htbp]
\centering
\caption{Comparison between PolarSym and SymPoint V2 on the benchmark.}
\label{tab:sympoint_compare}
\begin{tabular}{l l c c c c}
\toprule
Method & GPU / Batch & PQ (\%) & RQ (\%) & SQ (\%) & mIoU \\
\midrule
SymPoint V2 (Reproduction) & 2 / 8 & 85.42 & 89.92 & 94.98 & 66.56 \\
PolarSym (Ours) & 2 / 6 & \textbf{87.15} & \textbf{91.46} & \textbf{95.28} & \textbf{70.87} \\
Relative Improvement & & $+1.73$ & $+1.54$ & $+0.30$ & $+4.31$ \\
\bottomrule
\end{tabular}
\end{table}

\subsection{Ablation Experiments}

To investigate the contributions of individual components, we progressively introduced the directional and radial branches into the baseline network.

As shown in Table \ref{tab:ablation_components}, the baseline performance corresponds to the model without the PolarSym decoder module. Introducing the SF-RoPE module alone increased the PQ from 85.42\% to 86.83\% and improved the mIoU by 2.60\%. This result demonstrates that explicit directional modeling effectively establishes global geometric consistency and provides reliable directional cues for attention computation.

Further introduction of the RDB module elevated the performance to 87.15\% PQ and 70.87\% mIoU. Although the absolute gain in PQ is relatively modest, the RDB consistently improved all evaluation metrics. This observation indicates that directional modeling and radial modeling provide complementary geometric information. The directional branch determines the global structural orientation, while the radial branch refines long-range geometric correspondences.

We further evaluated the robustness of the proposed geometric modeling strategy under different initialization settings. Table \ref{tab:different_settings} compares PolarSym geometric modeling with a conventional combination of standard axial RoPE and linear distance bias.

When intentionally constructing weaker baselines via suboptimal initialization, the conventional geometric modeling achieved significant performance improvements: the baseline model's PQ increased by 4.2 percentage points, and the final model's PQ increased by 4.42 percentage points. Experimental results demonstrate that explicit geometric modeling facilitates Transformer optimization. Nevertheless, the proposed PolarSym consistently achieves significantly higher final performance. This result confirms that independent modeling of direction and distance outperforms direct concatenation of conventional positional encoding with linear distance bias.

\begin{table}[htbp]
\centering
\caption{Performance comparison of PolarSym components.}
\label{tab:ablation_components}
\begin{tabular}{lcccc}
\toprule
Setting & PQ (\%) & RQ (\%) & SQ (\%) & mIoU \\
\midrule
Baseline & 85.420 & 89.927 & 94.988 & 66.562 \\
+ SF-RoPE & 86.825 & 91.208 & 95.195 & 69.159 \\
+ RDB & 87.155 & 91.468 & 95.284 & 70.870 \\
\bottomrule
\end{tabular}
\end{table}

\begin{table}[htbp]
\centering
\caption{Performance comparison of PolarSym under different experimental settings.}
\label{tab:different_settings}
\begin{tabular}{l l c c c}
\toprule
Setting & Geometry Implementation & Baseline PQ & Final PQ & Relative Gain ($\Delta$) \\
\midrule
Weak Baseline & Standard Axial + Linear Dist & 81.019 & 83.57 & $+2.55\%$ \\
Strong Baseline & PolarSym (SF-RoPE + RDB) & 85.21 & 87.15 & $+1.94\%$ \\
\bottomrule
\end{tabular}
\end{table}

\subsection{Convergence Analysis}

\begin{figure}[htbp]
\centering
\includegraphics[width=0.50\linewidth]{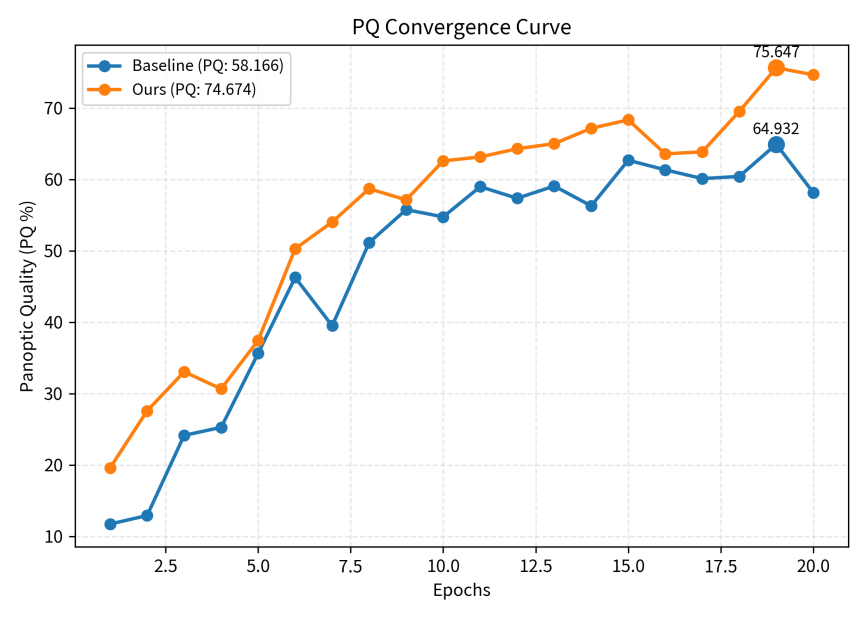}
\caption{PolarSym convergence process.}
\label{fig:Fig6}
\end{figure}

Beyond final performance, PolarSym demonstrates significantly faster convergence during training. Figure \ref{fig:Fig6} illustrates the PQ curves of the reproduced baseline and PolarSym.

In the first 20 epochs of training, PolarSym consistently outperforms the baseline in convergence speed. By the 20th epoch, PolarSym achieves 74.6\% PQ, whereas the baseline reaches only 58.1\% PQ. This acceleration primarily stems from explicit geometric modeling. Since the directional branch participates in attention computation from the outset, the network establishes global structural correspondences prior to learning fine-grained semantic representations. Subsequently, the radial branch progressively refines long-range geometric relationships, enabling stable optimization and improved convergence. The faster convergence rate further demonstrates that PolarSym enhances optimization efficiency rather than merely increasing model capacity.

\subsection{Qualitative Results}

\begin{figure}[htbp]
\centering
\includegraphics[width=0.50\linewidth]{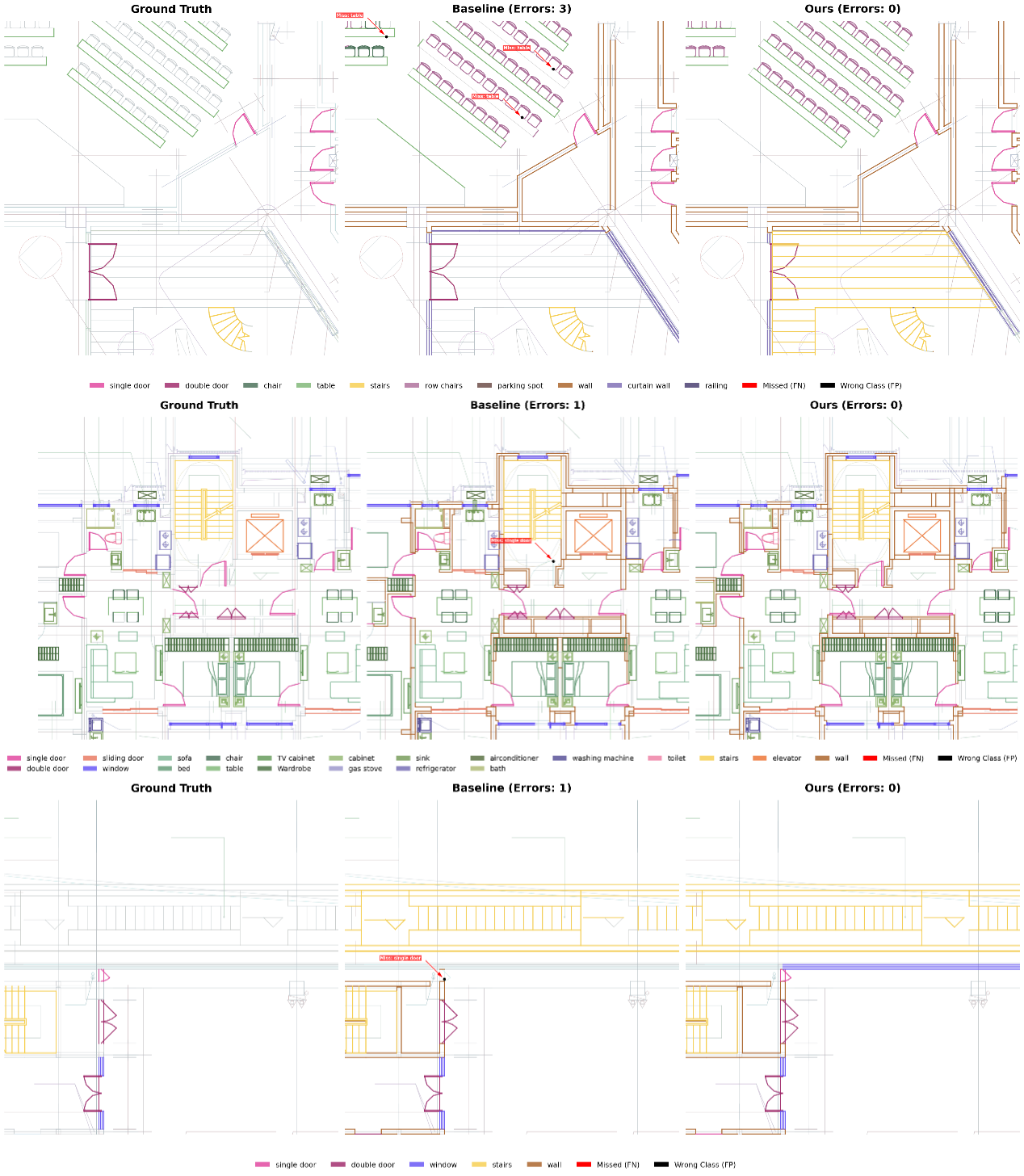}
\caption{Qualitative comparison of PolarSym and SymPoint V2 benchmarks on several challenging planar graph examples.}
\label{fig:Fig7}
\end{figure}

Figure \ref{fig:Fig7} illustrates the qualitative comparison between PolarSym and the reproduced SymPoint V2 benchmark on several challenging planar examples. Compared to the benchmark, PolarSym generates more complete structural predictions while reducing false positives in complex architectural regions. Notably, the proposed method achieves more accurate predictions for spatially separated yet geometrically correlated symmetric walls, doors, and furniture.

The visual improvements align with the previously described quantitative analysis. By explicitly modeling directional consistency and radial correspondence, PolarSym establishes more reliable long-range attention mechanisms, resulting in more coherent architectural layouts and reduced structural inconsistencies.

\section{Conclusion}

This work addresses the issue of transformer attention mechanisms lacking explicit geometric modeling in CAD floor plan parsing by proposing PolarSym, a polar coordinate geometry attention framework. Unlike existing methods that uniformly encode spatial information into positional representations, PolarSym explicitly decouples architectural geometric relationships into two complementary factors—direction and distance—and models them separately using SF-RoPE and RDB, respectively. Additionally, a lightweight dynamic gating mechanism is introduced to enable collaborative modeling of semantic features and geometric priors while preserving the original transformer architecture unchanged.

Extensive experimental results demonstrate the effectiveness of the proposed method. Under a unified constrained training environment, PolarSym outperforms the reproduced SymPoint V2 baseline across multiple metrics including PQ, RQ, SQ, and mIoU. Furthermore, it exhibits faster convergence speed and a more stable optimization process. Ablation studies further confirm the complementary nature of direction modeling and distance modeling: the former establishes global directional consistency, while the latter optimizes long-range structural correspondences. Together, they enhance the transformer’s geometric perception capability for building symmetry structures. Due to its lightweight extension to attention computation, PolarSym can be seamlessly integrated into existing transformer architectures without requiring modifications to the backbone network.

Although PolarSym achieves promising performance in CAD floor plan parsing, its geometric modeling remains primarily constrained to 2D planar structures. Future work will explore more generalizable geometric attention mechanisms, including unified modeling for 3D building models, cross-modal building data, and more complex spatial topological relationships, to further advance the transformer’s geometric reasoning capability in building scene understanding.

\bibliographystyle{alpha}
\bibliography{references}

\end{document}